\documentclass[letterpaper]{article} % DO NOT CHANGE THIS
\usepackage[submission]{aaai25}  % DO NOT CHANGE THIS
\usepackage{times}  % DO NOT CHANGE THIS
\usepackage{helvet}  % DO NOT CHANGE THIS
\usepackage{courier}  % DO NOT CHANGE THIS
\usepackage[hyphens]{url}  % DO NOT CHANGE THIS
\usepackage{graphicx} % DO NOT CHANGE THIS
\usepackage{natbib}  % DO NOT CHANGE THIS AND DO NOT ADD ANY OPTIONS TO IT
\usepackage{caption} % DO NOT CHANGE THIS AND DO NOT ADD ANY OPTIONS TO IT
\usepackage{algorithm}
\usepackage{algorithmic}

\usepackage{amsmath}
\usepackage{xspace}
\usepackage{amssymb}
\usepackage{multirow}
\usepackage{enumitem}
\usepackage{makecell}
\usepackage{url}            % simple URL typesetting
\usepackage{booktabs}       % professional-quality tables
\usepackage{amsfonts}       % blackboard math symbols
\usepackage{nicefrac}       % compact symbols for 1/2, etc.
\usepackage{microtype}      % microtypography
\usepackage{xcolor}         % colors
\usepackage{soul}

\usepackage{newfloat}
\usepackage{listings}
\DeclareCaptionStyle{ruled}{labelfont=normalfont,labelsep=colon,strut=off} % DO NOT CHANGE THIS
\floatstyle{ruled}
\newfloat{listing}{tb}{lst}{}
\floatname{listing}{Listing}
\newtheorem{definition}{Definition}

\newcommand{\paratitle}[1]{\vspace{0.8ex}\noindent \textbf{#1}}

\title{Toward Embodied AGI: A Review of Embodied AI and the Road Ahead}

\author{
  Yequan Wang ~~~~Aixin Sun\\
  \textnormal{tshwangyequan@gmail.com\\axsun@ntu.edu.sg}
}

\begin{document}

\maketitle

\begin{abstract}

Artificial General Intelligence (AGI) is often envisioned as inherently embodied. With recent advances in robotics and foundational AI models, we stand at the threshold of a new era—one marked by increasingly generalized embodied AI systems. This paper contributes to the discourse by introducing a systematic taxonomy of Embodied AGI spanning five levels (L1–L5). We review existing research and challenges at the foundational stages (L1–L2) and outline the key components required to achieve higher-level capabilities (L3–L5). Building on these insights and existing technologies, we propose a conceptual framework for an L3+ robotic brain, offering both a technical outlook and a foundation for future exploration.

\end{abstract}

%%%%%%%%%%%%%%%%%%%%%%%%%%%%%%%%%%%%%%%%%%%%%%%%%%%
\section{Introduction}
\label{sec:intro}
%%%%%%%%%%%%%%%%%%%%%%%%%%%%%%%%%%%%%%%%%%%%%%%%%%%

Artificial General Intelligence (AGI) has attracted considerable attention in recent years \cite{sparks}. Meanwhile, Embodied AI has also seen rapid advancement \cite{embodied}. It is widely recognized that Embodied AI is either an essential pathway to achieving AGI—reflecting the indispensable role of the human body in cognition \cite{body,schema}—or should even form part of AGI’s definition itself \cite{how_far_agi, path}. Rather than further examining the relationship between Embodied AI and AGI, we focus on \textit{Embodied AGI}, starting from the current literature on \textit{Embodied AI}, and explore how far it is from being truly \textit{humanoid and general}. We propose a pragmatic definition of Embodied AGI as follows:

\begin{definition}[Embodied AGI]
Embodied AGI is a form of Embodied AI that demonstrates human-like interaction capabilities and can successfully perform diverse, open-ended real-world tasks at a human-level proficiency.
\end{definition}

\begin{figure*}
    \centering
    \includegraphics[scale=0.45]{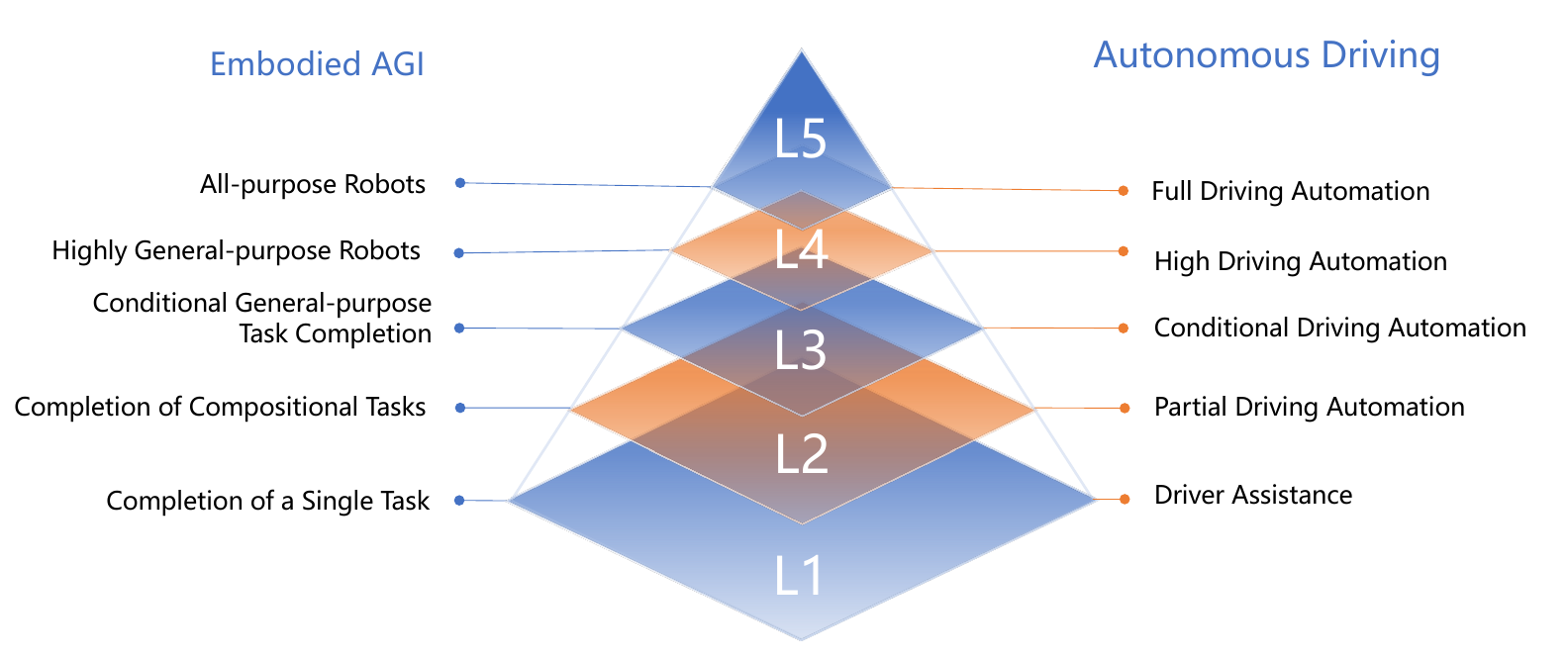}
    \caption{\textbf{Roadmap of the five levels of Embodied AGI, inspired by the established levels of autonomous driving.}}
    \label{fig:levels}
\end{figure*}

In this definition, Embodied AGI is framed as the intersection of AGI and Embodied AI, with an emphasis on human-like settings. To benchmark progress toward this goal, it is necessary to establish a set of criteria that clarifies the ultimate objective, assesses current capabilities, defines intermediate stages, and identifies key challenges and potential accelerators. Inspired by the levels of autonomous driving \cite{drivinglevels}, we introduce a five-level roadmap for Embodied AGI (Section \ref{sec:roadmap} and Figure~\ref{fig:levels}), ranging from Level 1 (L1)-assisting with a limited set of elementary tasks, to Level 5 (L5)-independently performing open-ended tasks with humanoid behaviors.

\begin{figure*}
    \centering
    \includegraphics[scale=0.6]{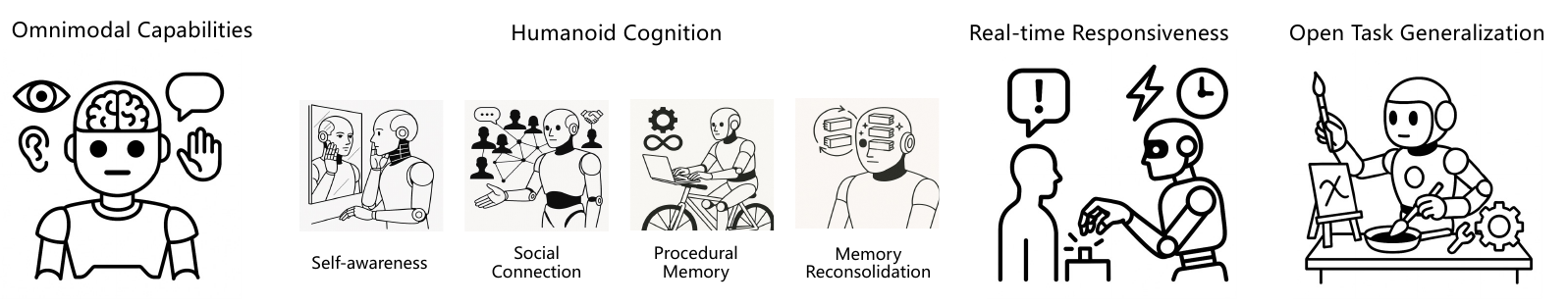}
    \caption{\textbf{Illustration of four basic constituents of Embodied AGI.} For a detailed discussion of the four subdomains of humanoid cognition, please see Section \ref{sec:l3l5}.}
    \label{fig:cognitive}
\end{figure*}

We assess the capabilities of embodied AI in four core dimensions:
(1) \textit{omnimodal capabilities}: the ability to  process a full spectrum of information modalities;
(2) \textit{humanoid cognitive abilities}: for nuance social comprehension and human-like learning mechanisms, including self-awareness, social connection understanding, procedural memory, and memory reconsolidation, as detailed in Section \ref{sec:l3l5};
(3) \textit{real-time responsiveness}: the capability to conduct swift, accurate actions and duplex interaction; 
(4) \textit{generalization}: the capability to adapt to open-ended environment and real-world tasks. The four dimensions are illustrated in Figure \ref{fig:cognitive}.

Based on the proposed five-level taxonomy and the four core dimensions of capability,we contextualize both recent developments and future directions. Recent developments in foundation models and embodied learning algorithms are briefly reviewed in Section~\ref{sec:l1l2}, evaluating their current maturity. Our analysis reveals that significant gaps remain across all four dimensions in reaching L3+ Embodied AGI, placing the current state of Embodied AI development between Levels 1 and 2 (L1–L2). In Section~\ref{sec:l3l5}, we identify the requirements across these four dimensions necessary for reaching Level 3 and beyond.

\begin{table*}[thbp]
\centering
\caption{\textbf{Definitions of L1 $\sim$ L5 explained.} We summarize the requirements of different capabilities.}
\scalebox{0.88}
{
\begin{tabular}{c|cccc|c|c}
\toprule
Level   & Modalities   & Humanoid & Real-time  & Generalization & Body \& Control & Autonomous Driving Analogy \\
\midrule
L1 & Partial & No & No & Unseen Environments & Robust & Simple Tasks (e.g., Speed Control) \\
L2 & Partial & No & No & Similar Tasks & + Responsive & Combined Tasks (e.g., Parking)  \\
L3 & Full & No & Partial & Limited Task Types & + Sensory-complete & Complex Tasks under Human Monitoring\\
L4 & Full & Partial & Yes & Open Tasks & + Accurate & Broad Tasks, Humanoid Accuracy\\
L5 & Full & Yes & Yes & Open Tasks & + Safe & All Tasks, no Human Intervention \\
\bottomrule
\end{tabular}
}
\label{tab:levels_def_drive}
\end{table*}

We observe that existing model architectures and widely adopted frameworks—such as Large Language Models (LLMs) \cite{gpt3, llama3}, Vision-Language Models (VLMs) \cite{llava, pixtral}, Vision-Language-Action (VLA) models \cite{openvla,RT-2}, and recent omnimodal approaches \cite{gemini1.5,qwen-omni}—fall short of meeting the requirements for L3+ multimodal processing and precise real-time action execution. Furthermore, prevailing learning paradigms, including supervised and reinforcement learning \cite{ppo}, remain inadequate for acquiring human-like behaviors and achieving robust generalization.

To help address these challenges, we propose a conceptual framework for L3+ Embodied AI learning in Section~\ref{sec:conceptual}. It comprises two key components: (i) a model architecture for an advanced robotic agent, and (ii) an integrated set of learning algorithms designed to satisfy the core requirements: \textit{omnimodal} processing, \textit{humanoid} cognitive abilities, \textit{real-time} responsiveness, and robust \textit{generalization}. The proposed architecture and algorithms are illustrative examples drawn from current research and may be replaced by future innovations, provided they fulfill the same foundational objectives.

%%%%%%%%%%%%%%%%%%%%%%%%%%%%%%%%%%%%%%%%%%%%%%%%%%%
\section{L1 $\sim$ L5: Roadmap to Embodied AGI}
\label{sec:roadmap}
%%%%%%%%%%%%%%%%%%%%%%%%%%%%%%%%%%%%%%%%%%%%%%%%%%%
Inspired by the widely accepted five levels of autonomous driving \cite{drivinglevels} and recent discussions on levels of AGI \cite{how_far_agi, morris2023levels}, we propose a five-stage roadmap toward Embodied AGI (L1–L5). This roadmap, summarized in Figure~\ref{fig:levels} and detailed in Table~\ref{tab:levels_def_drive}, defines each level based on four core dimensions (Figure \ref{fig:cognitive}): \textit{modalities}, \textit{humanoid} cognitive abilities, \textit{real-time} responsiveness, and \textit{generalization} capability. We also briefly outline hardware and locomotion/manipulation requirements \cite{gu2025humanoid} in Table~\ref{tab:levels_def_drive}, along with analogies to autonomous driving.

\paratitle{L1: Single-task completion.} At this level, embodied intelligent agents (e.g., robots) reliably perform a single, well-defined task—such as object grasping—that is useful in industrial or everyday settings. While they may exhibit limited generalization to novel conditions (e.g., changes in lighting or layout), their functionality remains confined to a specific task domain. Each single task typically requires a purpose-built robot with minimal versatility, while complex goals must be manually decomposed by humans into simpler sub-tasks. This stage is analogous to early large language models focused solely on machine translation (e.g., Chinese-to-English translation) \cite{vaswani2017attention}, or an L1 autonomous driving agent that handles isolated tasks like speed control and lane keeping. The robot’s physical body, at this level, must be sufficiently robust to support the execution of its target task.

\paratitle{L2: Compositional task completion.} At Level 2, robots can handle compositional tasks by decomposing high-level human instructions into sequences of simpler actions (e.g., grasping followed by cutting). Their broader skill set makes them more versatile than L1 robots and reduces the need for human intervention. However, their capabilities remain bounded to predefined tasks and skill libraries, with limited generalization beyond those domains. In the LLM literature, this corresponds to a multilingual machine translation system—able to translate between many language pairs, but still confined to the translation domain. In the autonomous driving literature, this is similar to handling combined tasks with explicit decomposition logic (e.g., parking), while more complex intellectual tasks (e.g., traffic jam navigation) remain out of reach. In addition to physical robustness, the robotic body at this level must be responsive enough to support longer and more complex action sequences.

\paratitle{L3: Conditional general-purpose task completion.} At Level 3, robots are capable of handling a wide range of task categories (e.g., grasping versus dancing), demonstrating conditional generalization across tasks, environments, and human instructions. They exhibit substantial real-time responsiveness, dynamically adapting to environmental changes or updated directives. However, while versatile and capable of multi-tasking, their performance on entirely novel or open-ended tasks is not yet reliable. Thus, L3 represents an early stage of general-purpose embodied intelligence. Supporting this level requires a robotic body with comprehensive sensory input (e.g., vision, audition, optionally touch and proprioception) and corresponding output modalities. In the context of LLMs, this stage loosely resembles pre-trained foundation models (e.g., BERT, GPT-3, LLaMA 2) equipped with multitask fine-tuning or capable of few-shot prompting \cite{bert,gpt3,llama-2}. In autonomous driving, this corresponds to solving complex tasks such as long-term highway driving and traffic navigation, albeit under human monitoring.

\paratitle{L4: Highly general-purpose robots.} Starting from L4, robots exhibit robust generalization to a broad range of unseen tasks, marking true general-purpose capability. Such robots effectively internalize scientific laws and physical world models \cite{world-model}, enabling accurate predictions and decision-making. In addition to real-time processing, they possess strong multimodal comprehension and reasoning abilities (e.g., across language, audio, and vision), ensuring sophisticated communication and interaction with humans. The robotic body is expected to be more flexible and accurate to match these advanced capabilities. For LLM analogy, L4 robots can be roughly considered as the general-purpose LLMs equipped with strong reasoning capabilities, such as o1 \cite{o1} and DeepSeek-R1 \cite{dpskr1}. For autonomous driving analogy, L4 resembles solving most driving tasks at near-human accuracy, but a minimum level of human intervention is still involved.

\paratitle{L5: All-purpose robots.}
L5 represents the ultimate goal of Embodied AGI: developing genuinely all-purpose robotic agents capable of meeting a wide spectrum of human daily needs. These robots integrate a deep understanding of physical laws and human emotional and social dynamics, processing all modalities seamlessly in real-time. They exhibit distinctly human-like cognitive behaviors, including self-awareness, social connection understanding, procedural memory, and memory reconsolidation (Section \ref{sec:l3l5}). At this level, the robotic body should incorporate safety mechanism to prevent potential dangerous intentions from being executed. For LLM analogy, L5 corresponds to a still-emerging stage of textual AGI. In autonomous driving, it reflects a complete understanding of nuanced human needs in driving scenarios, thus fully eliminating the need for human intervention.

%%%%%%%%%%%%%%%%%%%%%%%%%%%%%%%%%%%%%%%%%%%%%%%%%%%
\section{L1 $\sim$ L2: \textit{Status Quo} and Challenges}
\label{sec:l1l2}
%%%%%%%%%%%%%%%%%%%%%%%%%%%%%%%%%%%%%%%%%%%%%%%%%%%

We begin with a brief literature review to assess the current status of embodied AI. Two mainstream approaches dominate the field: \textit{end-to-end} and \textit{plan-and-act}. End-to-end methods typically leverage Visual-Language-Action (VLA) models, directly processing visual and textual inputs to generate actions via next-token prediction \cite{RT-2,openvla} or diffusion-based methods \cite{ALOHA}. Conversely, plan-and-act approaches first utilize Visual-Language Models (VLMs) or Large Language Models (LLMs) to interpret multimodal inputs and then perform high-level planning and task decomposition, generating intermediate control signals such as executable code \cite{Voxposer}, function calls \cite{SMART-LLM}, or verbal instructions \cite{vila}. Some hybrid methods integrate both paradigms through latent-space planning \cite{agibot}. The notable success of LLMs \cite{GPT-4,llama3} has significantly influenced foundational model development within embodied AI, promoting large-scale pre-training strategies using real-world and synthetic datasets to enhance generalization \cite{GR-2, pi05, vpp}.

\textit{What level have we reached?} Our review suggests that the capabilities required for L1 Embodied AGI are already fully or partially met by existing models. Many can reliably complete single tasks with demonstrated robustness to previously unseen environments and conditions. For example, GraspVLA \cite{graspvla} successfully generalizes grasping across various lighting conditions, backgrounds, distractions, and object heights. Yet, it remains specialized in grasping tasks and does not generalize beyond this domain. State-of-the-art robotic systems, such as Helix\footnote{\url{https://www.figure.ai/news/helix}}, not only display robust generalization within specific task types (e.g., picking diverse objects) but can also handle a wide range of dexterous indoor tasks. Such robots approach L2-level proficiency by decomposing complex human instructions into executable sub-tasks and solving them either independently or through coordinated two-bot systems. 

Advancing to Level-3 (L3) requires handling substantially different task categories and exhibiting robust real-time responsiveness. Recent works, like $\pi_{0.5}$ \cite{pi05}, partly address diverse task categories through combined pre-training (e.g., mobile and non-mobile tasks), yet their applications still largely focus on environmental generalization rather than genuine task diversity. Thus, we conclude that current Embodied AI capabilities are positioned between Levels 1 and 2 (L1–L2).

We identify four critical challenges hindering the progression of embodied AI to L3 and beyond, covering each of the four dimensions:

\paratitle{Lack of comprehensive joint-modal capabilities.} Predominant models (e.g., VLA) typically incorporate only vision and textual language inputs, generating outputs solely in the action space. True embodied intelligence necessitates full-spectrum multimodal perception (e.g., understanding human speech with emotion and sentiment; listening to environmental audio inputs from microphone devices in addition to text console and imagery camera) and multimodal responses, including real-time acoustic speech feedback. The absence of such modalities  not only severely restricts the versatility of embodied agents in application, but also prevents them from a thorough understanding of the physical world.

\begin{figure*}
    \centering
    \includegraphics[scale=0.8]{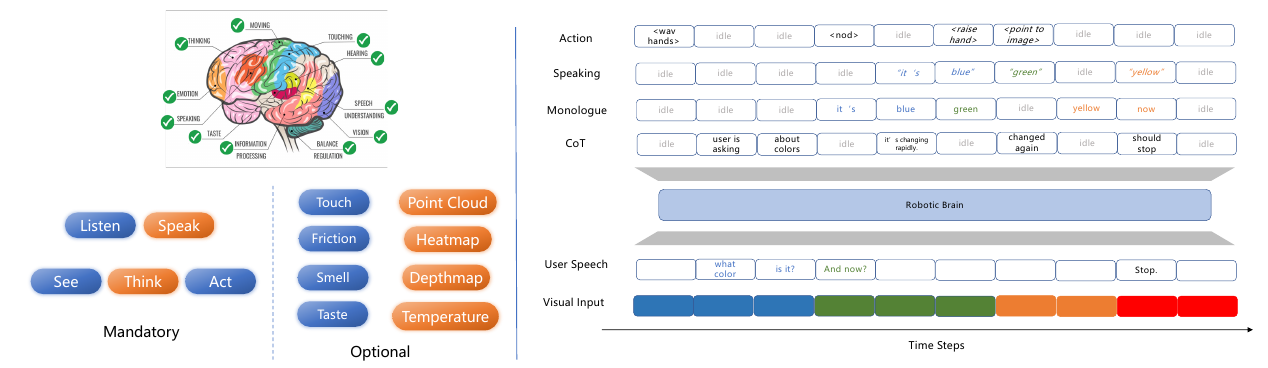}
    \caption{\textbf{Omnimodal Capabilities and Illustrative Model Structure.}}
    \label{fig:conceptual-model}
\end{figure*}

\paratitle{Insufficient humanoid cognition.} Existing robots primarily focus on achieving task-specific manipulations without adequately addressing higher-level intellectual interactions or nuanced communications. Fully capable embodied agents must excel in reasoning and conversational intelligence, akin to sophisticated chatbots \cite{o1,dpskr1}, and demonstrate alignment with human preferences and ethical values \cite{instructgpt}. Ultimately, for L5, agents should exhibit distinctly humanoid cognitive behaviors and sophisticated social comprehension, which remain far beyond the reach for current learning paradigms including unsupervised, supervised, and reinforcement learning.

\paratitle{Limited real-time responsiveness.} Most current embodied AI systems operate in a semi-duplex manner: receiving and processing instructions fully before acting, making them struggle for dynamic environments where conditions or instructions change rapidly. This limitation significantly impedes real-world deployment.

\paratitle{Restricted generalization.}
As mentioned above, recent embodied AI models have made substantial progress in generalizing across diverse environments. However, it is worth noticing that there are still a wide range of cross-environment generation scenarios that current models struggle to handle, a typical example being the invariance to spatial transformations (e.g., camera angles) \cite{spatial}. These issues must be addressed to reach higher levels. More importantly, inter-task generalization is still underdeveloped but essential for achieving true general-purpose capabilities (L3+).

%%%%%%%%%%%%%%%%%%%%%%%%%%%%%%%%%%%%%%%%%%%%%%%%%%%
\section{L3 $\sim$ L5: Key Constituents}
\label{sec:l3l5}
%%%%%%%%%%%%%%%%%%%%%%%%%%%%%%%%%%%%%%%%%%%%%%%%%%%
In this section, we delve into the essential constituents of L3+ Embodied AGI derived from their definitions. We analyze recent advancements achieved by the research community, examine challenges that current methods face in reaching higher levels, and propose potential technical paths and design choices to bridge these gaps.

\paratitle{Omnimodal capabilities.}
A fundamental requirement of L3–L5 Embodied AGI is their ``general-purpose'' nature, achievable only through comprehensive omnimodal capabilities extending beyond vision and language. This is because real-world applications frequently demand an understanding of auditory cues, human speech nuances, tactile feedback, thermal perception, and more. Moreover, for L4 and beyond, mastery of these additional modalities becomes critical for acquiring and internalizing knowledge of physical laws, which is potentially the foundation of true generalization capability. 

While bimodal foundational models such as visual-language \cite{GR-2} and audio-language models \cite{glm-voice} have been extensively explored, and tri-modal models (e.g., vision-language-audio) have garnered considerable interest recently \cite{gemini1.5, qwen-omni}, incorporating additional modalities like actions and environmental sensing for embodied agents remains largely uncharted. Moreover, current models face two critical challenges:(1) modality conflicts, which impose high demands on model capacity \cite{multiscaling}; and (2) cascading errors and alignment issues arising from modality-specific modules and heterogeneous data distributions \cite{cambrian}. To address these issues, future models necessitate (1) \textit{parallel} understanding-inference-generation architectures (L3+) to effectively control the time complexity imposed by model capacity, and (2) more advanced multimodal pre-training paradigms (particularly for L4+) that improves the collaboration of modality-specific modules or inherently supports multimodal comprehension. 

\paratitle{Humanoid cognitive behaviors.} Human-like cognitive behaviors are essential across all levels (L1–L5) because (1) mimicking essential learning mechanisms of human neural brain \cite{neuralbrain} potentially enhances the capability of embodied agents, and (2) a humanoid understanding of self and social connections improves the quality of human-robot interaction. Ultimately, L4+ robots should seamlessly integrate into human daily life by recognizing individual users, understanding emotional contexts, and even developing a sense of identity and social bonds \cite{sumers2023cognitive}. We consider four capabilities being the core of achieving humanoid cognition (Figure \ref{fig:cognitive}): 

\begin{itemize}
    \item \textit{Self-awareness.} As supported by cognitive science \cite{self-aware-cog} and philosophy \cite{self-aware-phil}, self-awareness is the foundation of higher cognitive functions. A self-aware agent can understand its identity, temporal continuity, and objectives with greater nuance \cite{self-aware-llm}. This awareness should be lifelong, dynamic, and stateful—rather than statically encoded in a system prompt, as in most current LLMs.
    \item \textit{Social connection understanding.} Understanding the relationships between oneself and other humans or robots—as well as relationships among others—is a higher-order cognitive capability. Such awareness helps an AI system comprehend its roles, responsibilities, and character \cite{socialbench}, enhancing its ability to participate in role-based interactions, especially in L4+ settings. Like self-awareness, a true social connection understanding should also be lifelong, dynamic, and stateful \cite{lifelong}.
    \item \textit{Procedural memory.} Humans maintain an extendable memory of incrementally learned skills, known as procedural memory \cite{procedural}. In AI, this is related to overcoming domain shifts \cite{gem} and addressing catastrophic forgetting \cite{overcoming-cata}. Agents equipped with procedural memory can accumulate and refine skills over time.
    \item \textit{Memory reconsolidation.} Most current machine learning systems produce static model checkpoints after training, disallowing further learning during deployment. In contrast, humans continuously evaluate the salience of new information and update their knowledge based on time, context, and experience—an ability known as memory reconsolidation \cite{reconsolidation}. For embodied AGI, this capability is critical not only to reduce re-training overhead, but also to enable long-term adaptation and intelligence evolution.
\end{itemize}

Most of aforementioned cognition behaviors are closely connected to lifelong learning. Although recent studies have emphasized long-context learning \cite{needle}, efforts have primarily focused on extending the context window \cite{flashattention} and optimizing positional encodings \cite{xpos}. In contrast, lifelong learning entails an unbounded temporal scope, wherein a model continuously updates its internal states and memory representations within its parameters, rather than relying on external caches \cite{lifelong}. Human-like understanding of identity, social dynamics, and emotional contexts emerges through lifelong experiential learning supported by long-term memory. Humanoid models should therefore adopt similar lifelong learning paradigms, maintaining continuously updated internal representations of self, knowledge, and the external environment, through active, ongoing interactions.

\begin{figure*}
    \centering
    \includegraphics[scale=0.55]{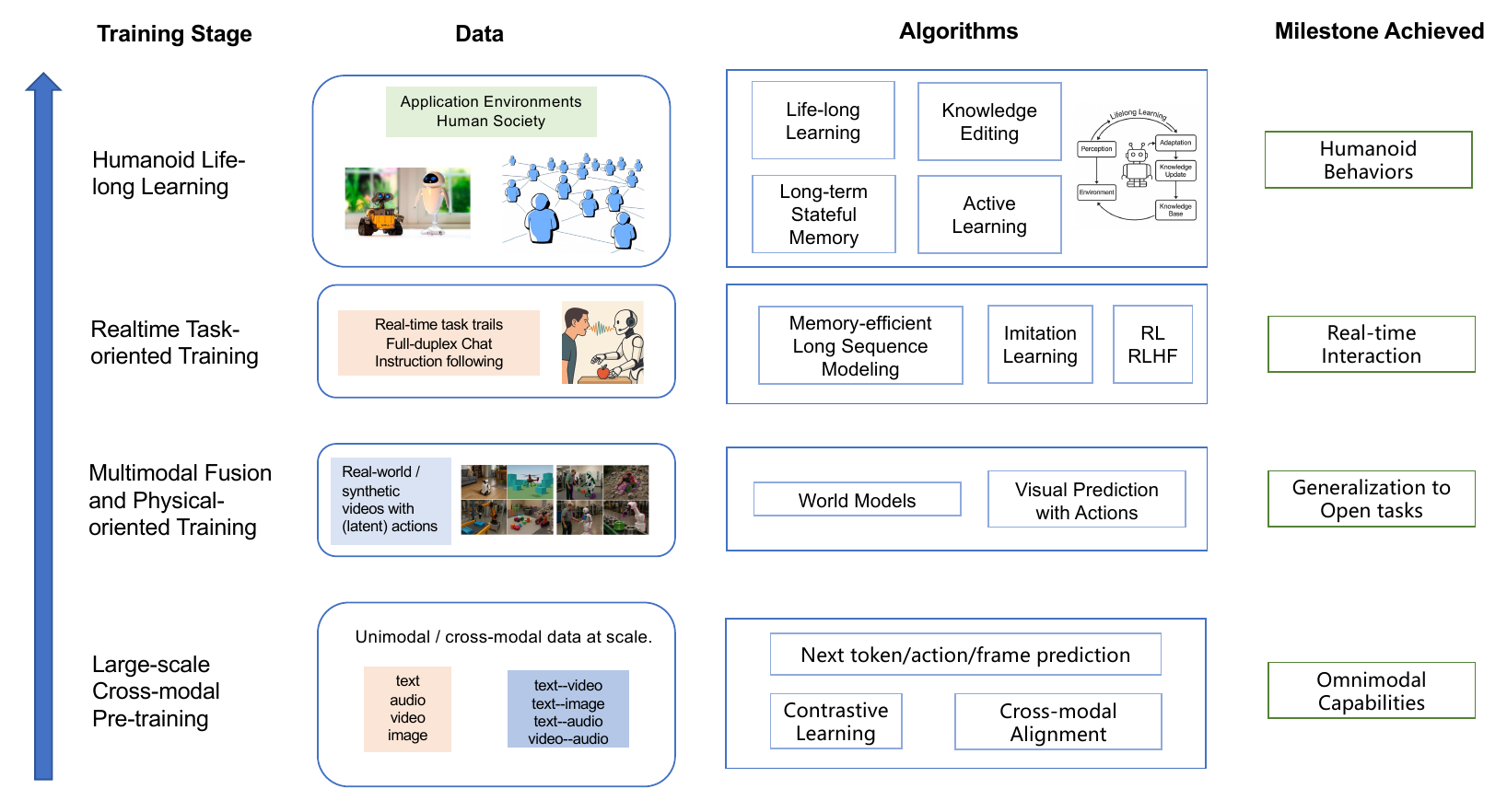}
    \caption{\textbf{Conceptual Framework: Illustrative Training Paradigms.}}
    \label{fig:conceptual-paradigm}
\end{figure*}

\paratitle{Real-time interaction.} Real-time responsiveness is essential across nearly all Embodied AI applications, especially for general-purpose agents at L3 and above, which must adapt to dynamic real-world environments and swiftly respond to rapidly-changing human instructions. Currently, real-time manipulation often imposes model-size constraints; for example, models such as GO-1 \cite{agibot} and $\pi_{0.5}$ \cite{pi05} employ VLA architectures restricted to under 5 billion parameters. Besides, real-time auditory and visual interactions are commonly implemented using Time Division Multiplexing (TDM) methods \cite{beyond,full-dup}. However, these approaches encounter scalability issues when incorporating additional modalities, as computational complexity increases quadratically with sequence length \cite{vaswani2017attention}. Engineering-oriented optimizations, such as those implemented in MiniCPM-o\footnote{\url{https://github.com/OpenBMB/MiniCPM-o}}, partially alleviate this bottleneck. Nevertheless, achieving L3+ real-time performance will require new paradigms specifically designed to support genuinely multiplexed, omnimodal processing.

\paratitle{Generalization to open-ended tasks.} As discussed in Section \ref{sec:l1l2}, current embodied AI models demonstrate notable generalization across varied environments but struggle to generalize effectively across diverse task categories. A central limitation that hinders widely-considered unsupervised or multitask pretraining approaches from solving the problem of task generation is their insufficient internalization of \textit{physical world laws}, which restricts their ability to accurately predict the outcomes of virtual or imagined actions. As a result, models often overfit to task-specific cues rather than uncovering underlying generalizable principles. Developing training objectives beyond simple imitation or generation—such as predictive modeling of physical interactions or causal reasoning—could significantly enhance inter-task generalization and better prepare embodied agents for open-ended, heterogeneous tasks.

%%%%%%%%%%%%%%%%%%%%%%%%%%%%%%%%%%%%%%%%%%%%%%%%%%%
\section{A Conceptual Framework for L3+ Robots}
\label{sec:conceptual}
%%%%%%%%%%%%%%%%%%%%%%%%%%%%%%%%%%%%%%%%%%%%%%%%%%%

In this section, we propose a conceptual framework specifically designed to meet the requirements for developing L3+ Embodied AGI, as outlined in Section \ref{sec:roadmap}. This framework is composed of an omnimodal model structure and a corresponding training paradigm that potentially supports the emergence of L3+ capabilities.

\subsection{Model Structure}

As discussed in Section \ref{sec:l3l5}, essential characteristics of an L3–L5 embodied AI model structure include comprehensive modality integration and native real-time interactions. Ideally, at each timestep $t+1$, the model should generate responses conditioned on all prior information observed at timesteps $0 \ldots t$. Specifically, the model jointly processes multimodal input streams, such as simultaneous audio and video, and generates multimodal outputs, including action sequences, continuous speech, internal monologues, and chain-of-thought reasoning, etc.:

\begin{align}
    y_{a_1}^{t+1}, &y_{a_2}^{t+1},...y_{a_n}^{t+1} = f_{\theta}(x_{b_1}^{0\sim t}, x_{b_2}^{0\sim t},...x_{b_m}^{0\sim t}).\\
    a_i &\in \scalebox{0.9}{\{thoughts, speech, action, mobile...\}};\\
    b_j &\in \scalebox{0.9}{\{text, audio, image, video, heatmap...\}};
\end{align}

An illustrative architecture is presented in Figure \ref{fig:conceptual-model}. The proposed structure supports omnimodal streaming input and output, facilitating rapid response to dynamic real-world conditions, such as changing human instructions, interruptions, environmental perturbations, and immediate feedback from previous actions. A bi-modal prototype case of such architecture is RQ-Transformer \cite{moshi}.

\subsection{Training Paradigm}

An illustrative training paradigm is depicted in Figure \ref{fig:conceptual-paradigm}, detailing the required data, learning algorithms, and milestone targets at each stage. The algorithms referenced are drawn from the \textit{current AI literature} and may be replaced by future innovations serving similar objectives. The motivation and components for this paradigm are detailed as follows:

\paratitle{Multimodal training from scratch.} We advocate training inherently multimodal models from scratch to facilitate deep cross-modal alignment and omnimodal understanding. A crucial research direction involves developing effective training stages and dataset arrangements to maximize cross-modal interactions and facilitate joint-modal comprehension.

\paratitle{Lifelong learning.} Inspired by human cognitive behaviors, we propose moving beyond the traditional ``pre-train → fine-tune → deploy'' paradigm toward lifelong, continuous learning frameworks \cite{lifelong, zheng2025lifelong}, integrating related methodologies such as active learning \cite{activellm} and knowledge editing \cite{knowledge-editing} for multimodal embodied agents.

\paratitle{Physical-oriented training.} To improve generalization across open-ended task scenarios essential for higher-level Embodied AGI, we propose to explore training paradigms oriented toward physical-world understanding. These approaches should leverage unsupervised or synthetic data at scale and incorporate explicit or implicit actions within the learning objective, allowing models to internalize causal effects and physical laws. Promising directions include outcome-prediction frameworks driven by fine-grained actions \cite{vpp}, and the expansion of generalized World Models \cite{world-model, navigation} to cover a broader domain of tasks and interaction dynamics.

\section{Conclusion and Future Challenges}

In this paper, we present a comprehensive review of the development of Embodied AGI by establishing a five-level taxonomy as a roadmap, benchmarking current progress, identifying critical capability gaps, and proposing a conceptual framework. We argue that this roadmap remains relevant in the long term, though advancements in robotic hardware, infrastructure, and machine learning may lead to the evolution, modification, or replacement of the proposed framework as an implementation strategy. 

Our discussion is grounded in the premise that Embodied AGI should demonstrate human-like intellectual behaviors. Consequently, future challenges will not only include technical barriers but also ethical and safety considerations, as well as broader societal implications—particularly concerning the dynamics of collaboration and relationships among humans, robots, and human-robot collectives. 

We hope this paper contributes valuable insights and stimulates meaningful discussion toward a more informed and constructive future for embodied general intelligence.

\bibliography{custom}

\end{document}